\documentclass[runningheads]{llncs}
\usepackage[T1]{fontenc}
\usepackage{graphicx}
\usepackage{notation}
\usepackage[capitalise]{cleveref}
\begin{document}

\sloppy
\title{A Connection Between Learning to Reject and Bhattacharyya Divergences}
%
%
\author{Alexander Soen\inst{1,2}\orcidID{0000-0002-2440-4814}}
\authorrunning{A. Soen}
%
\institute{Australian National University, Canberra, Australia \and
	RIKEN AIP, Tokyo, Japan
	}
\maketitle              
\begin{abstract}
Learning to reject provide a learning paradigm which allows for our models to abstain from making predictions. 
One way to learn the rejector is to learn an ideal marginal distribution (\wrt the input domain) --- which characterizes a hypothetical best marginal distribution --- and compares it to the true marginal distribution via a density ratio.
In this paper, we consider learning a joint ideal distribution over both inputs and labels; and develop a link between rejection and thresholding different statistical divergences.
We further find that when one considers a variant of the log-loss, the rejector obtained by considering the joint ideal distribution corresponds to the thresholding of the skewed Bhattacharyya divergence between class-probabilities.
This is in contrast to the marginal case --- that is equivalent to a typical characterization of optimal rejection, Chow's Rule --- which corresponds to a thresholding of the Kullback-Leibler divergence. In general, we find that rejecting via a Bhattacharyya divergence is less aggressive than Chow's Rule.

\end{abstract}
\section{Introduction}

Learning to reject~\cite{cOO,zdhRW,cdmLW,nchsOT,msCE} is a learning setting in ML which allows for models to abstain from making a prediction --- where abstained inputs can eventually handled by a human expert or another model downstream.
From a formal perspective, we consider the classification task between an input domain $\mathcal{X}$ and a space of labels $\mathcal{Y} \defeq [L]$. We assume that examples for the classification task are produced via a joint distribution $\meas{P} \in \triangle(\mathcal{X} \times \mathcal{Y})$, where $\triangle(\mathcal{Z})$ denote the probability simplex for the space $\mathcal{Z}$. In addition, we define a space related to the labels $\tilde{Y}$ in which our models will produce outputs to, \eg, the space of logits $\tilde{\mathcal{Y}}$ versus the space of labels $\mathcal{Y}$.
Now, given a prediction model $h \colon \mathcal{X} \to \tilde{\mathcal{Y}}$, we seek to find a combined model $h^{\rej} \colon \mathcal{X} \to \tilde{\mathcal{Y}} \cup \{ \rejtoken \}$ such that

\begin{equation}
	\label{eq:combined}
	h^{\rej}(x) 
    \;
    =
    \;
        h(x) 
        \quad 
        \textrm{if } 
        \quad
        r(x) = 0 \quad \textrm{otherwise } 
        \quad
        \rejtoken
\end{equation}
where $\rejtoken$ is a rejection token signifying abstention and $ r \colon \mathcal{X} \to \{ 0, 1 \} $ is a binary function which controls when inputs are abstained, \ie, when $r(x) = 1$ the input $x$ is being abstained for the model $h$.

From an optimization perspective, there are various choices which can be made on which objects we learn in the learning to reject pipeline. In this paper, we will focus on the \emph{post-hoc} rejection setting in which the model $h$ is considered fixed and we only wish to optimize for a rejecting function $r$~\cite{mmzPR,njmrkPH,jgmnrkWD}.
To learn a rejector $r \colon \mathcal{X} \to \{ 0, 1 \}$ in the post-hoc case, one can consider the following optimization problem~\cite{fpvOS,hpvmdML}:
\begin{equation}
	\label{eq:obj}
	\min_{r \colon \mathcal{X} \to \{ 0, 1 \}}
	\quad
	\expect_{(\X, \Y) \sim \meas{P}}\left[
		(1 - r(\X)) \cdot \ell(\Y, h(\X))
		\right]
	+
	c \cdot \meas{P}[r(\X) = 1],
\end{equation}
where $c \geq 0$ corresponds to a cost of rejection and $ \ell \colon \mathcal{Y} \times \tilde{\mathcal{Y}} \rightarrow \mathbb{R}_{\geq 0} $ is a loss function to punish miss-prediction, \eg, log-loss / cross-entropy.


One common approach in the learning to reject literature is to consider the zero-one loss function $\ell = \ell_{\zo}$ in \cref{eq:obj} and then seek to minimize surrogate loss functions of the zero-one rejection objective~\cite{ccfggansGC,cdmBW,mmzRH}. Recently, an alternative perspective has been proposed which involves the optimization of an \emph{ideal distribution} $\meas{Q}_{\rm x} \in \triangle(\mathcal{X})$ in which a rejection mechanism can be derived via the thresholding of a density ratio $\dratio \colon \mathcal{X} \to \mathbb{R}_{\geq 0}$~\cite{shsnRV}. Concretely, the ideal distribution $\meas{Q}$ can be obtained via the optimization
\begin{equation}
	\label{eq:rejobj}
	\mathop{\arg\min}_{\meas{Q}_{\rm x} \in \triangle(\mathcal{X})}
	\quad
	\expect_{\X \sim \meas{Q}_{\rm x}} \left[
		\expect_{\Y \sim \pos^\star(\X)} \left[
			\ell(\Y, h(\X))
			\right]
		\right]
	+
	\lambda \cdot \diss(\meas{Q}_{\rm x} \Mid \meas{P}_{\rm x}),
\end{equation}
where $\pos^{\star}(x) \in  \triangle(\mathcal{Y})$ corresponds to the Bayes posterior / ground-truth class-probability function of the prediction task and $\meas{P}_{\rm x}$ corresponds to the marginal distribution of the prediction task, \ie, $\meas{P}(x, y) = \meas{P}_{\rm x}(x) \cdot \pos_y^{\star}(x)$. The weighted latter term corresponds to a dissimilarity function $\diss \colon \triangle(\mathcal{X}) \times \triangle(\mathcal{X}) \rightarrow \mathbb{R}_{\geq 0}$, where we assume that $\lambda \geq 0$.

The final rejector $r \colon \mathcal{X} \rightarrow \{ 0, 1 \}$ is obtained via the thresholding
\begin{equation}
	\label{eq:drrej}
	\dratio(x) = \frac{\dmeas{Q}_{\rm x}}{\dmeas{P}_{\rm x}}(x); \quad\quad
	r^{\dr}(x; \tau) =  \llbracket \dratio(x) \leq \tau \rrbracket,
\end{equation}
where $\llbracket p \rrbracket$ are Iverson brackets~\cite{kTN} which evaluate to $1$ when the predicate $p$ is true and evaluate to $0$ when $p$ is false.

From an intuitive perspective, the shift from finding a binary function $r$ directly (in \cref{eq:obj}) to find a distribution $\meas{Q}_{\rm x}$ (in \cref{eq:rejobj}) can be seen as optimizing for the induced reweighting of ``$(1-r(x)) \cdot \meas{P}_{\rm x}(x)$''. Instead of penalizing frequent rejection via $\meas{P}[r(\X) = 1]$, we penalize deviation of $\meas{Q}_{\rm x}$ from the ground-truth $\meas{P}$ via a dissimilarity function $\diss(\meas{Q}_{\rm x} \Mid \meas{P}_{\rm x})$.

It has been shown that if one takes the dissimilarity function in \cref{eq:rejobj} to be the Kullback-Leibler divergence (KL divergence), then the well-known criteria for optimal rejection --- \emph{Chow's Rule}~\cite{cOO} --- is recovered~\cite[Theorem 4.2]{shsnRV}.
However, if one deviates from the density ratio derived from utilizing the KL divergence in \cref{eq:drrej}, then what is considered optimal will differ from Chow's Rule.

In this paper, we consider a modification to the density ratio rejection framework and seek to learn a \emph{joint ideal distribution} (in $\triangle(\mathcal{X} \times \mathcal{Y})$) instead of an ideal marginal distribution in $\meas{Q} \in \triangle(\mathcal{X})$.
A rejector can then be derived from this joint ideal distribution via first marginalizing the labels $\mathcal{Y}$ and then thresholding a density ratio (similar to \cref{eq:drrej}).
When one considers a variant of the log-loss $\ell = \tilde{\ell}_{\log}$, we find that optimal rejector derived from this joint ideal distribution actually corresponds to the thresholding of skewed Bhattacharyya divergences~\cite{bOA} between estimated and ground-truth class-probability functions.


\section{Chow's Rule and Optimal Rejection}
\label{sec:rej}

To be more concrete, we assume that $\tilde{\mathcal{Y}} = \mathbb{R}^L$, \ie, the space of logit values. That is the codomain of the models $h$ correspond to $\mathbb{R}^L$ and one can map a model's output to the simplex $\triangle(\mathcal{Y})$ via $\pos_{y}(x) \propto \exp(h_{y}(x)) \in [0, 1]$, \ie, $\pos(x) \in \triangle(\mathcal{Y})$. This mapping to the simplex corresponds to the common softmax function. Using this probability mapping, we can also provide a predictor $f(x) = \arg\max_{y \in \mathcal{Y}} \pos_y(x)$.

Given the optimization problem of \cref{eq:obj}, it is possible to characterize Chow's Rule, an optimal notion of rejection. 

\begin{theorem}[Chow's Rule]
    \label{thm:chow}
    The optimal rejector $r^\star$ of \cref{eq:obj} is given by
    \begin{equation}
        \label{eq:chow}
        r^\star(x; c) = \left \llbracket
            \expect_{\Y \sim \pos^\star(x)} \left[
            \ell(\Y, h(x))
            \right] \geq c
        \right \rrbracket.
    \end{equation}
\end{theorem}

This result follows almost immediately from factorizing the objective in \cref{eq:obj} \wrt $r(x)$ and minimizing point-wise for the binary outcomes of $r$. The inner expectation in \cref{eq:chow} is often called the \emph{conditional risk} of the loss $\ell$~\cite{rwID}.


\begin{example}
    \label{ex:log_loss}
    Suppose we consider the log-loss / cross-entropy loss $\ell_{\log}(y, h(x)) = -\log \pos_y(x)$. From the definition of the cross-entropy function, the optimal rejector can be written as a function of the KL divergence between the estimated class probabilities $\pos$ and the Bayes posterior $\pos^\star$. Indeed, defining $\kl(\mu \Mid \nu) = \expect_{\mu}[\log \frac{\dmeas{}\mu}{\dmeas{}\nu}] $, we have that
    $r_{\log}^\star(x; c) = \llbracket \kl(\pos^\star(x) \Mid \pos(x)) \geq c'(x) \rrbracket $, where $c'(x) = c - \ent(\pos^\star(x))$ with $\ent(\mu) = \expect_{\mu}[- \log \mu]$ corresponding to Shannon entropy. Notice that the threshold $c'(x)$ of the KL divergence is not constant.
\end{example}

So far we have considered (optimal) rejectors derived from the classically considered learning to reject objective, \cref{eq:obj}. The optimal rejector determined by the density ratio framework outlined by \cref{eq:rejobj,eq:drrej} can also be found in closed form when considering (differentiable) 
$f$-divergences~\cite{cOI} 
as the chosen dissimilarity function $ \diss(\cdot \Mid \cdot)$~\cite{shsnRV}. In this paper, we will focus on the case where we consider the canonical KL divergence.

\begin{theorem}[{\cite[Corollary 4.1]{shsnRV}}]
    \label{thm:ideal_dr}
    Consider the optimization problem in \cref{eq:rejobj} where $\diss = \kl$. In this case, the optimal density ratio $\dratio$ is given by
    \begin{equation}
        \label{eq:dratio}
        \dratio(x) 
        =
        \frac{1}{Z} \cdot \exp\left(
            - \frac{\expect_{\Y \sim \pos^\star(x)} \left[
            \ell(\Y, h(x))
            \right]
            }{\lambda}
        \right),
    \end{equation}
    where 
    \(
        Z = \expect_{\X \sim \meas{P}_{\rm x}}\left[
            \exp\left(
                - {\expect_{\Y \sim \pos^\star(\X)} \left[
                \ell(\Y, h(\X))
                \right]
                }/{\lambda}
            \right)
        \right].
    \)
\end{theorem}

The ideal distribution $\meas{Q}_{\rm x}$ can also be found with \cref{thm:ideal_dr} via $\meas{Q}_{\rm x} = \meas{P}_{\rm x} \cdot \dratio$. In other words, the ideal distribution when considering the KL divergence is an exponential reweighting of the true (marginal) data distribution $\meas{P}_{\rm}$.
In the sequel, we will denote $r^\dr(x; \tau) \defeq \llbracket \dratio(x) \geq \tau \rrbracket$ with $\dratio$ defined by \cref{eq:dratio}.

Via the monotonicity of $z \mapsto \exp(-z / \lambda)$ for $\lambda > 0$, a rejector $r^\dr$ obtained by thresholding $\dratio$ is equivalent to the optimal rejector $r^\star$ described by \cref{thm:chow}.

\begin{theorem}[{\cite[Theorem 4.2]{shsnRV}}]
    \label{cor:connect_chow}
    Let $\lambda > 0$. 
    For any cost $c \geq 0$, there exists a density ratio rejector $r^\dr$ that is equivalent to optimal rejection given by \cref{thm:chow}.
    That is, there exists a $\tau$ such that $r^\dr(\cdot; \tau) = r^\star(\cdot; c)$.
\end{theorem}
\section{Rejection through Joint Ideal Distributions}
\label{sec:joint}

The ideal distribution $\meas{Q}_{\rm x}$ found via the optimization problem of \cref{eq:rejobj} can intuitively be thought of as a hypothetical distribution in which the classifier $h$ performs best on. 
The ideal distributions described so far only correspond to distributions on the input space $\mathcal{X}$. In the following section, we outline how ideal distributions can be extended to joint distributions on $\mathcal{X} \times \mathcal{Y}$ and consider a `joint' variant of rejection via density ratios.

We first consider how one can learn a joint ideal distribution. To do so, we simply replace the optimization of the marginal in \cref{eq:rejobj} to optimization of a joint distribution:
\begin{equation}
    \label{eq:jointrejobj}
    \mathop{\arg\min}_{\meas{Q} \in \triangle(\mathcal{X} \times \mathcal{Y})}
    \quad
    \expect_{\X \sim \meas{Q}} \left[ 
        \ell(\Y, h(\X))
    \right]
    +
    \lambda \cdot \diss(\meas{Q} \Mid \meas{P}).
\end{equation}

A solution to \cref{eq:jointrejobj} provides a joint ideal distribution $\meas{Q} \in \triangle(\mathcal{X} \times \mathcal{Y})$. However, as we are interested in rejection, the dependence on $\mathcal{Y}$ makes it inadequate for use in test-time, \ie, if we want to evaluate the joint ideal distribution, we would need the label  $y$ that we are predicting a-priori.
As such, to consider a density ratio rejector which is not a function of $\mathcal{Y}$, we first marginalize the joint ideal distribution to yield the following rejector:
\begin{equation}
    \meas{Q}_{\rm j}(x) = \int_{\mathcal{Y}} \meas{Q}(x, y) \; \dmeas{}y; \quad \dratio_{\rm j}(x) = \frac{\dmeas{Q}_{\rm j}}{\dmeas{P}_{\rm x}}(x); \quad r_{\rm j}^\dr(x; \tau) = \llbracket \dratio_{\rm j}(x) \leq \tau \rrbracket.
\end{equation}

We call $\dratio_{\rm j}$ and $r_{\rm j}^\dr$ the joint ideal density ratio and the joint ideal density ratio rejector, respectively.
Similar to the marginal ideal density ratio, the joint ideal density ratio can be expressed in an analytical form when considering the KL divergence as a measure of dissimilarity.
\begin{theorem}
    \label{thm:joint_ideal_dr}
    Consider the optimization problem in \cref{eq:jointrejobj} where $\diss = \kl$. In this case, the optimal joint density ratio $\dratio_{\rm j}$ is given by
    \begin{equation}
        \label{eq:jointdratio}
        \dratio_{\rm j}(x) 
        =
        \frac{1}{Z_{\rm j}} \cdot \expect_{\Y \sim \pos^\star(x)} \left[
            \exp\left(
                - \frac{
                \ell(\Y, h(x))
                }{\lambda}
            \right)
        \right],
    \end{equation}
    where 
    \(
        Z_{\rm j} = \expect_{(\X, \Y) \sim \meas{P}}\left[
            \exp\left(
                - {\ell(\Y, h(\X))
                }/{\lambda}
            \right)
        \right].
    \)
\end{theorem}

One might notice that the difference between the marginal in \cref{thm:ideal_dr} and the joint density ratio obtained in \cref{thm:joint_ideal_dr} is where the condition expectation (\wrt $\pos^\star$) is being evaluated. This difference can also be interpreted as a result from applying Jensen's inequality. The inequality results in the following relation between the marginal and joint density ratios.
\begin{lemma}
    \label{lem:relate_rej}
    For the KL divergence ideal density ratios, we have that
    \begin{equation}
        \label{eq:dratio_relate}
        Z \cdot \dratio(x) \leq Z_{\rm j} \cdot \dratio_{\rm j}(x),
    \end{equation}
    where $\frac{Z}{Z_{\rm {j}}} \leq 1$.
    Furthermore, for $ \tau' = \frac{Z}{Z_{\rm j}} \cdot \tau $, the corresponding rejectors have that
    \begin{equation}
        \label{eq:rej_relate}
        r^\dr(x; \tau) \geq r_{\rm j}^\dr(x; \tau').
    \end{equation}
\end{lemma}

From \cref{lem:relate_rej}, we have that with an appropriate scaling of the threshold decision $\tau$, the joint ideal density ratio rejector is less aggressive. That is whenever $r_{\rm j}^\dr(x; \tau)$ rejects, 
$r^\dr(x; \tau')$ must reject. It should also be noted that the inequality in \cref{eq:dratio_relate,eq:rej_relate} solely comes from Jensen's inequality. As such, \cref{eq:rej_relate} has equality iff the loss of the model $\ell(h(x), y)$ is constant. However, this would imply the model performance of $h$ is constant (\wrt the loss) and thus rejection either reject all or none examples.
Hence, in general, the joint ideal density ratio rejector is less aggressive.
\section{Connection to the Bhattacharyya Divergence}
\label{sec:bhatta}

In the following section, we develop a connection between the joint density ratio rejector outlined in \cref{thm:joint_ideal_dr} and the family of skewed Bhattacharyya divergences~\cite{bOA,nTC}. To do so, we consider a slight modification to the log-loss $\ell_{\log}$.
To establish this connection we first define the Bhattacharyya divergence.

\begin{definition}
	\label{def:bhatta}
	The skewed Bhattacharyya Divergence $\bhatta_{\beta}(\meas{Q} \Mid \meas{P})$ for $\beta \in (0, 1)$ is defined by
	\begin{equation}
		\bhatta_{\beta}(\meas{Q} \Mid \meas{P})
		\defeq
		- \log \bcoeff_{\beta}(\meas{Q} \Mid \meas{P})
	\end{equation}
	where $\bcoeff_{\beta}(\meas{Q} \Mid \meas{P})$ is the the skewed Bhattacharyya coefficient defined by
	\begin{equation}
		\bcoeff_{\beta}(\meas{Q} \Mid \meas{P})
		\defeq
		\int_{z \in \mathcal{Z}} \meas{Q}(z)^{\beta} \meas{P}(z)^{1 - \beta} \dmeas{}z.
	\end{equation}
\end{definition}

To connect rejection via a joint density ratio rejector to Bhattacharyya, we first define the following variant of the log-loss $\ell_{\log}$ discussed in \cref{ex:log_loss}:
\begin{equation}
	\label{eq:modified_log}
	\tilde{\ell}_{\log}(y, h(x))
	\defeq
	- \log \pos_{y}(x) + \log \pos^\star_{y}(x)
	=
	- \log \frac{\pos_{y}(x)}{\pos^\star_{y}(x)}.
\end{equation}
It should be noted that $\tilde{\ell}_{\log}$ depends on $x$ implicitly in its arguments.

\begin{remark}
	Although \cref{eq:modified_log} appears to be an odd choice of loss function, it actually corresponds to learning a rejector by comparing the loss of $h$ to the ground-truth. Indeed, considering the original rejection objective in \cref{eq:obj} with $\ell = \tilde{\ell}_{\log}$ one gets the following simplification:
	\begin{align}
		 & \expect_{(\X, \Y) \sim \meas{P}}\left[
		(1 - r(\X)) \cdot
		\ell_{\log}(\Y, h(\X))
		+
		r(\X) \cdot
		\ell_{\log}(\Y, h^\star(\X))
		\right] \nonumber                         \\
		 & \quad \quad \quad \quad
		+
		c \cdot \meas{P}[r(\X) = 1]
		+
		\expect_{\X \sim \meas{P}_{\rm x}}[
			\ent(\pos^\star(\X))
		],
		\label{eq:tilde_log_loss}
	\end{align}
	where $h^\star$ corresponds to the optimal model obtained by utilizing the Bayes optimal posterior $\pos^\star$.

	Intuitively, \cref{eq:tilde_log_loss} essentially defers to the optimal model $h^\star$ whenever the rejector $r(x) = 1$ would reject. That is, instead of paying no cost when we would reject (when considering just $\ell = \ell_{\log}$), we pay for the potential uncertainty of $\pos^\star$. This is equivalent to considering a \emph{model cascade} setting: instead of learning a rejection function just for $h$, we want to learn a router $r$ which determines whether examples $x$ should go to model $h$ or $h^\star$~\cite{jgmnrkWD}. 
    \cref{eq:tilde_log_loss} can be interpreted as assuming that the second model corresponds to the Bayes posterior, which is similar in assumption to the distillation learning setting~\cite{hvdDT}.
\end{remark}

Now taking the modified log-loss $\tilde{\ell}_{\log}$, we establish a connection between the joint ideal density ratio rejector and the skewed Bhattacharyya divergences.

\begin{theorem}
	\label{thm:connect_bhatta}
	Let $\ell = \tilde{\ell}_{\log}$.
	When $\lambda > 1$, the optimal joint density ratio $\dratio_{\rm j}$ can be expressed as
	\begin{equation}
		\label{eq:bcoeff_rejector}
		\dratio_{\rm j}(x)
		=
		\frac{1}{Z_{\rm j}} \cdot \bcoeff_{1/\lambda}(\pos(x) \Mid \pos^\star(x)).
	\end{equation}
	Furthermore, for $\kappa_{\rm j} = - \log (Z_{\rm j} \cdot \tau)$,
	\begin{equation}
		\label{eq:bhatta_rejector}
		r^\dr_{\rm j}(x; \tau)
		=
		r^{\dr}_{\rm j}(x; \kappa_{\rm j})
		\defeq
		\llbracket
		\bhatta_{1 - \frac{1}{\lambda}} (\pos^\star(x) \Mid \pos(x)) \geq \kappa_{\rm j}
		\rrbracket.
	\end{equation}
\end{theorem}

In \cref{thm:connect_bhatta} --- with slight abuse of notation --- we introduce a reparameterization of the thresholding in the rejector to be a function of $\kappa_{\rm j}$ for simplicity.
It should be noted that the restriction of $\lambda > 1$ in \cref{thm:connect_bhatta} is only to ensure that the skew of the $\beta = 1 - (1 / \lambda)$ is a valid skew as per \cref{def:bhatta}. Despite this, the functional form will also hold for $\lambda > 0$.
One could also consider a rejector as a function of the Bhattacharyya coefficient $ \bcoeff_{1/\lambda}(\pos(x) \Mid \pos^\star(x))$. A benefit of using the Bhattacharyya coefficient for rejection is its bounded property $\bcoeff_{1/\lambda}(\pos(x) \Mid \pos^\star(x)) \in [0, 1]$ (a result of H\"{o}lder's inequality)~\cite{nRC}.

It should be noted that switching from a traditional log-loss $\ell_{\log}$ to the modified log-loss $\tilde{\ell}_{\log}$ still makes rejection via Chow's Rule (or an optimal marginal joint density ratio rejector) dependent on the KL divergence. The only difference is that instead of depending on a thresholding cost $c'(x)$ as a function of $\mathcal{X}$ (as per \cref{ex:log_loss}), the thresholding of the KL divergence becomes a constant.

\begin{corollary}
	\label{cor:connect_kl}
	Let $\ell = \tilde{\ell}_{\log}$. For $\kappa = - \lambda \cdot \log(Z \cdot \tau)$, the optimal marginal density ratio rejector is given by
	\begin{equation}
		r^\dr(x; \tau)
		=
		r^\dr(x; \kappa)
		\defeq
		\llbracket
		\kl(\pos^\star(x) \Mid \pos(x)) \geq \kappa
		\rrbracket.
	\end{equation}
\end{corollary}

Similar to \cref{thm:connect_bhatta}, in \cref{cor:connect_kl} we provide a parameterization by $\kappa$ for the rejector for simplicity.

Now comparing the density ratio rejectors obtained via \cref{thm:connect_bhatta} and \cref{cor:connect_kl}, it can clearly be seen that the difference in rejection is the type of divergence being utilized to threshold.
An interesting point of comparison is the role of $\lambda$ in each rejector. In the marginal case, $\lambda$ only changes the threshold point of rejection. In the joint case, $\lambda$ changes the skew of the Bhattacharyya divergence --- which changes the boundary an example is rejected, rather than just the threshold value --- that we consider to determine rejection.

There is a strong connection between the skewed Bhattacharyya divergences and the KL divergence. Indeed, one first notes the connection between Bhattacharyya divergences and R\'{e}nyi divergences~\cite{nRC}; where
the Bhattacharyya coefficient and divergence have been previously called the $\alpha$-R\'{e}nyi affinity and the unnormalized R\'{e}nyi divergence, respectively~\cite{gTM}.
Then we can exploit the monotonicity of $\alpha$ to connect to the KL divergences, in which the $(\alpha=1)$-R\'{e}nyi divergence corresponds to the KL-divergence~\cite[Theorem 3]{vhRD}.

\begin{lemma}
	\label{lem:bhatta_kl_connect}
	For $\lambda > 1$, we have that
	\begin{equation}
		\bhatta_{1 - \frac{1}{\lambda}} (\pos^\star(x) \Mid \pos(x))
		\leq
		\frac{1}{\lambda} \cdot \kl(\pos^\star(x) \Mid \pos(x)).
	\end{equation}
	As a result, for $\ell = \tilde{\ell}_{\log}$ and a divergence threshold $\kappa$,
	\begin{equation}
		r^\dr(x; \lambda \cdot \kappa) \geq r^\dr_{\rm j}(x; \kappa).
	\end{equation}
\end{lemma}

\cref{lem:bhatta_kl_connect} provides a divergence specific perspective of the connection between marginal and joint rejectors for the $\ell = \tilde{\ell}_{\log}$ case. In contrast to the more general connection in  \cref{lem:relate_rej}, the change in thresholding in \cref{lem:bhatta_kl_connect} is independent of the normalization constants $Z$ and $Z_{\rm j}$.




%
\begin{credits}
	\subsubsection{\ackname} 
        We thank Richard Nock and Hisham Husain for initial discussions on joint ideal distributions for rejection. We thank Frank Nielsen for their feedback on the manuscript and help finding key references.

\end{credits}
%
%
%
\bibliographystyle{splncs04}
\bibliography{mybibliography}
\newpage
\section*{Appendix}
\appendix
\section{Proof of \cref{thm:joint_ideal_dr}}

\begin{proof}
    The proof of \cref{thm:joint_ideal_dr} follows similarly to \cite[Corollary 4.1]{shsnRV}. We first find $\meas{Q}_{\rm j}$ via the method of Lagrange multipliers. Consider the functional
    \begin{align}
        \bar{\mathcal{J}}(\meas{Q}) 
        &\defeq \expect_{\X \sim \meas{Q}}[\ell(\Y, h(\X))] + \lambda \cdot \kl(\meas{Q} \Mid \meas{P}) \nonumber \\
        &= \expect_{\X \sim \meas{Q}}[\ell(\Y, h(\X))] + \lambda \cdot \expect_{\meas{Q}}\left[\log \frac{\dmeas{Q}}{\dmeas{P}}(\X, \Y) \right] \nonumber \\
        &= \expect_{\X \sim \meas{Q}}
        \left[
        \ell(\Y, h(\X)) + \lambda \cdot \log \frac{\dmeas{Q}}{\dmeas{P}}(\X, \Y) \right] \nonumber \\
        &= \iint_{\mathcal{X} \times \mathcal{Y}} \meas{Q}(x, y) \cdot \left(
        \ell(y, h(x)) + \lambda \cdot \log \frac{\meas{Q}(x, y)}{\meas{P}(x, y)} 
        \right)
        \dmeas{}(x, y).
        \label{eq:original_obj}
    \end{align}

    Now applying Lagrange multipliers to account for the simplex conditions on $\triangle(\mathcal{X} \times \mathcal{Y})$, we have
    \begin{align}
        {\mathcal{J}}(\meas{Q}; a, b) 
        &\defeq 
        \iint_{\mathcal{X} \times \mathcal{Y}}
        a(x, y) \cdot \meas{Q}(x, y)
        \dmeas{}(x, y)
        +
        b \cdot
        \left(
        \iint_{\mathcal{X} \times \mathcal{Y}}
        \meas{Q}(x, y)
        \dmeas{}(x, y)
        - 1
        \right) \nonumber \\
        & \quad +
        \iint_{\mathcal{X} \times \mathcal{Y}} \meas{Q}(x, y) \cdot \left(
        \ell(y, h(x)) + \lambda \cdot \log \frac{\meas{Q}(x, y)}{\meas{P}(x, y)} 
        \right)
        \dmeas{}(x, y).
        \label{eq:lagrange_obj}
    \end{align}
    Notice that \cref{eq:original_obj} is a convex functional of $\meas{Q}$ (it is the addition of a linear function and the convex KL-divergence). As such we have Lagrange duality.
    
    We observe that $\meas{Q}(x) > 0$ as otherwise \cref{eq:lagrange_obj} (the KL-divergence) will evaluate to $\infty$. Hence, by complementary slackness we must have $a(x) = 0$.
    Hence, by first order optimality conditions, the optimal $\meas{Q}^\star$ must satisfy the following condition on the functional derivative
    \begin{equation*}
        0
        =
        \frac{\dmeas{}}{\dmeas{}\delta}
        \mathcal{J}(\meas{Q} + \delta R; a, b) \bigg\vert_{\delta = 0},
    \end{equation*}
    for any $R \colon \mathcal{X} \times \mathcal{Y} \to \mathbb{R}$.

    Taking $a(x) = 0$ and letting $\delta > 0$, and denoting $\tilde{\meas{Q}}_{\delta} = \meas{Q}^{\star} + \delta \cdot R$, we have
    \begin{align*}
        &\frac{\dmeas{}}{\dmeas{}\delta}
        \mathcal{J}(\meas{Q} + \delta h; a, b) \\
        &= 
        \frac{\dmeas{}}{\dmeas{}\delta}
        \iint_{\mathcal{X} \times \mathcal{Y}} \tilde{\meas{Q}}_{\delta}(x, y) \cdot \left(
        b
        +
        \ell(\Y, h(\X)) + \lambda \cdot \log \frac{\tilde{\meas{Q}}_{\delta}(x, y)}{\meas{P}(x, y)} 
        \right)
        \dmeas{}(x, y) \\
        &= \iint_{\mathcal{X} \times \mathcal{Y}} 
        R(x, y)
        \cdot \left(
        b
        +
        \ell(y, h(x)) 
        +
        \lambda \cdot
        \log \frac{\tilde{\meas{Q}}_{\delta}(x, y)}{\meas{P}(x, y)}
        + \lambda
        \right)
        \dmeas{}(x, y).
    \end{align*}
    Taking $\delta \downarrow 0$ and setting the functional derivative to $0$, we have
    \begin{equation*}
        0 = 
        \iint_{\mathcal{X} \times \mathcal{Y}} 
        R(x, y)
        \cdot \left(
        b'
        +
        \ell(y, h(x)) 
        +
        \lambda \cdot 
        \log \frac{\meas{Q}^{\star}(x, y)}{\meas{P}(x, y)}
        \right)
        \dmeas{}(x, y),
    \end{equation*}
    where we let $b' = b + \lambda$.
    
    As this hold for all $R$, we must have point-wise that,
    \begin{align*}
        0 &= b' + \ell(y, h(x)) + \lambda \cdot \log \frac{\meas{Q}^{\star}(x, y)}{\meas{P}(x, y)} \\
        \iff \quad
        \meas{Q}^\star(x, y) &= \meas{P}(x, y) \cdot \exp\left( - \frac{\ell(y, h(x)) + b'}{\lambda}
        \right) \\
        \iff \quad
        \meas{Q}^\star(x, y) &= \exp\left(- \frac{b'}{\lambda}\right) \cdot \meas{P}(x, y) \cdot \exp\left( - \frac{\ell(y, h(x))}{\lambda}
        \right).
    \end{align*}
    As we must have $1 = \iint \meas{Q}^\star(x, y) \dmeas{}(x, y)$, we must have that
    \begin{equation*}
        \exp\left(- \frac{b'}{\lambda}\right) = \frac{1}{\expect_{(\X, \Y) \sim \meas{P}}\left[
            \exp\left( - \frac{\ell(y, h(x))}{\lambda}
        \right)
        \right]}
        = \frac{1}{Z_{\rm j}}.
    \end{equation*}

    Now we compute the proof by marginalizing $\meas{Q}^\star$,
    \begin{align*}
        \meas{Q}_{\rm j}(x) 
        &= \int_{\mathcal{Y}} \meas{Q}^\star(x, y) \dmeas{}y \\
        &= \frac{1}{Z_{\rm j}} \cdot \int_{\mathcal{Y}} \meas{P}(x, y) \cdot \exp\left( - \frac{\ell(y, h(x))}{\lambda}
        \right) \dmeas{}y \\
        &= \frac{1}{Z_{\rm j}} \cdot \meas{P}_{\rm x} \cdot \int_{\mathcal{Y}} \pos_y^\star(x) \cdot
        \exp\left( - \frac{\ell(y, h(x))}{\lambda}
        \right) \dmeas{}y \\
        &= \frac{1}{Z_{\rm j}} \cdot \meas{P}_{\rm x}(x) \cdot \expect_{\Y \sim \pos^\star(x)} \left[
            \exp\left(
                - \frac{
                \ell(\Y, h(x))
                }{\lambda}
            \right)
        \right],
    \end{align*}
    The proof is completed by computing the density ratio $\dratio_{\rm j} = \dmeas{Q}_{\rm j} / \dmeas{P}_{\rm x}$.
\end{proof}
\section{Proof of \cref{lem:relate_rej}}

\begin{proof}
    The proof follows from applications of Jensen's inequality. 
    First notice that $Z \leq Z_{\rm j}$ immediately from Jensen's inequalty.
    
    Furthermore, we have
    \begin{align*}
        Z \cdot \dratio(x)
        &= 
        \exp\left(
            - \frac{\expect_{\Y \sim \pos^\star(x)} \left[
            \ell(\Y, h(x))
            \right]
            }{\lambda}
        \right) \\
        & \leq 
        \expect_{\Y \sim \pos^\star(x)} \left[
        \exp\left(
            - \frac{
            \ell(\Y, h(x))
            }{\lambda}
        \right)
        \right] \\
        &= Z_{\rm j} \cdot \dratio_{\rm j}(x).
    \end{align*}

    Now to relate rejectors, we similarly note
    \begin{align*}
        r^\dr_{\rm j}(x; \tau)
        &= \llbracket \dratio_{\rm j}(x) \leq \tau \rrbracket \\
        &= \llbracket Z_{\rm j} \cdot \dratio_{\rm j}(x) \leq Z_{\rm j} \cdot \tau \rrbracket \\
        &\leq \llbracket Z \cdot \dratio(x) \leq Z_{\rm j} \cdot \tau \rrbracket \\
        &\leq \left\llbracket \dratio(x) \leq \frac{Z_{\rm j}}{Z} \cdot \tau \right\rrbracket \\
        &= r^\dr\left(x; \frac{Z_{\rm j}}{Z} \cdot \tau \right).
    \end{align*}
    Re-parameterizing $\tau$ yields the desired result.
\end{proof}
\section{Proof of \cref{thm:connect_bhatta}}

\begin{proof}
    We first note that with $\tilde{\ell}_{\log}$, we have
    \begin{align*}
        \exp\left( - \frac{\tilde{\ell}_{\log}(y, h(x))}{\lambda} \right)
        &= 
        \exp\left( \frac{\log \pos_{y}(x) - \log \pos^\star_{y}(x) }{\lambda} \right) \\
        &= 
        \exp\left( 
        \frac{1}{\lambda} \cdot \log \frac{\pos_{y}(x)}{\pos^\star_{y}(x)}
        \right) \\
        &= 
        \left(\frac{\pos_{y}(x)}{\pos^\star_{y}(x)} \right)^{1 / \lambda}.
    \end{align*}
    Now, utilizing $\dratio_{\rm j}$ from \cref{thm:joint_ideal_dr}, we have
    \begin{align*}
        \dratio_{\rm j}(x) 
        &= \frac{1}{Z_{\rm j}} \int_{y \in \mathcal{Y}} \pos^\star_{y}(x) \cdot 
        \left(\frac{\pos_{y}(x)}{\pos^\star_{y}(x)} \right)^{1 / \lambda} \dmeas{}y \\
        &= \frac{1}{Z_{\rm j}} \int_{y \in \mathcal{Y}} \pos^\star_{y}(x)^{1 - \frac{1}{\lambda}} \cdot 
        \pos_{y}(x)^{1 / \lambda} \dmeas{}y \\
        &= \frac{1}{Z_{\rm j}} \cdot \bcoeff_{1/\lambda}(\pos(x) \Mid \pos^\star(x)).
    \end{align*}

    Notice that by a re-parameterization of $\lambda$
    \begin{equation*}
        \bcoeff_{1/\lambda}(\pos(x) \Mid \pos^\star(x))
        =
        \bcoeff_{1 - \frac{1}{\lambda}}(\pos^\star(x) \Mid \pos(x))
        =
        \bcoeff_{\frac{\lambda - 1}{\lambda}}(\pos^\star(x) \Mid \pos(x)).
    \end{equation*}

    Now, we consider the rejector
    \begin{align*}
        r^\dr_{\rm j}(x; \tau)
        &= \left \llbracket 
            \frac{1}{Z_{\rm j}}
            \cdot
            \bcoeff_{\frac{\lambda - 1}{\lambda}}(\pos^\star(x) \Mid \pos(x)) \leq \tau
        \right \rrbracket \\
        &= \left \llbracket 
            \bcoeff_{\frac{\lambda - 1}{\lambda}}(\pos^\star(x) \Mid \pos(x)) \leq Z_{\rm j} \cdot \tau
        \right \rrbracket \\
        &= \left \llbracket 
            - \log \bcoeff_{\frac{\lambda - 1}{\lambda}}(\pos^\star(x) \Mid \pos(x)) \geq -\log(Z_{\rm j} \cdot \tau)
        \right \rrbracket \\
        &= \left \llbracket 
            \bhatta_{\frac{\lambda - 1}{\lambda}}(\pos^\star(x) \Mid \pos(x)) \geq -\log(Z_{\rm j} \cdot \tau)
        \right \rrbracket.
    \end{align*}
    As required.
\end{proof}
\section{Proof of \cref{cor:connect_kl}}

\begin{proof}
    First note the following relation:
    \begin{align*}
        \expect_{\Y \sim \pos^\star(x)} \left[
            \tilde{\ell}_{\log}(h(x), \Y)
        \right]
        &= \expect_{\Y \sim \pos^\star(x)} \left[
            - \log \pos_{\Y}(x) + \log \pos^\star_{\Y}(x)
        \right] \\
        &= \kl(\pos^\star(x) \Mid \pos(x)).
    \end{align*}

    Hence, by \cref{thm:ideal_dr}, we have that
    \begin{align*}
        \dratio(x) &= 
        \frac{1}{Z} \cdot \exp\left(
            - \frac{\expect_{\Y \sim \pos^\star(x)} \left[
            \ell(h(x), \Y)
            \right]
            }{\lambda}
        \right) \\
        &= \frac{1}{Z} \cdot \exp\left(
        - \frac{1}{\lambda}
        \cdot \kl(\pos^\star(x) \Mid \pos(x))
        \right).
    \end{align*}
    The corresponding rejector then is given by
    \begin{align*}
        r^\dr(x; \tau)
        &= \left \llbracket
        \frac{1}{Z} \cdot \exp\left(
        - \frac{1}{\lambda}
        \cdot \kl(\pos^\star(x) \Mid \pos(x))
        \right)
        \leq \tau
        \right \rrbracket \\
        &= \left \llbracket
        - \frac{1}{\lambda}
        \cdot \kl(\pos^\star(x) \Mid \pos(x))
        \leq \log (Z \cdot \tau)
        \right \rrbracket \\
        &= \left \llbracket
        \kl(\pos^\star(x) \Mid \pos(x))
        \geq \lambda \cdot \log (Z \cdot \tau)
        \right \rrbracket.
    \end{align*}
    As required.
\end{proof}

\section{Proof of \cref{lem:bhatta_kl_connect}}

\begin{proof}
    Let us denote the $\alpha$-R\'{e}nyi Divergence by $\renyi_{\alpha}(\cdot \Mid \cdot)$. The Bhattacharyya divergence can be written in terms on the R\'{e}nyi divergence as follows~\cite[Equation 5]{nRC}:
    \begin{equation}
        \bhatta_{\beta}(\meas{P} \Mid \meas{Q}) = (1 - \beta) \cdot \renyi_{\beta}(\meas{P} \Mid \meas{Q}).
    \end{equation}

    Now from \cite[Theorem 3]{vhRD} $\renyi_{\alpha}$ is non-decreasing in $\alpha$ and the $(\alpha = 1)$-R\'{e}nyi divergence corresponds to the KL-divergence \cite[Theorem 5]{vhRD}. Hence,
    \begin{align*}
        \bhatta_{\frac{\lambda - 1}{\lambda}} (\pos^\star(x) \Mid \pos(x))
        &= \left(1 - \frac{\lambda - 1}{\lambda} \right) \cdot \renyi_{\frac{\lambda - 1}{\lambda}}(\pos^\star(x) \Mid \pos(x)) \\
        &= \frac{1}{\lambda} \cdot \renyi_{\frac{\lambda - 1}{\lambda}}(\pos^\star(x) \Mid \pos(x)) \\
        &\leq \frac{1}{\lambda} \cdot \kl(\pos^\star(x) \Mid \pos(x)),
    \end{align*}
    noting that $(\lambda - 1) / \lambda < 1$ for $\lambda > 1$.

    Now the relation between rejectors follow immediately from applying the inequality between divergences:
    \begin{align*}
        r^\dr_{\rm j}(x; \kappa)
        &= \llbracket
        \bhatta_{\frac{\lambda - 1}{\lambda}} (\pos^\star(x) \Mid \pos(x)) 
        \geq \kappa
        \rrbracket \\
        &\leq 
        \left\llbracket
        \frac{1}{\lambda}
        \cdot
        \kl(\pos^\star(x) \Mid \pos(x)) 
        \geq \kappa
        \right\rrbracket \\
        &=
        r^\dr(x; \lambda \cdot \kappa).
    \end{align*}
    As required.
\end{proof}
\end{document}